\documentclass{svproc}
\usepackage{amsmath}
\usepackage{algorithm}  
\usepackage{algorithmicx}  
\usepackage{algpseudocode}  
\usepackage{array}
\usepackage{graphicx}
\usepackage{subfigure}
\usepackage{url}
\usepackage[misc]{ifsym}

\def\eg{\emph{e.g.,}}

\def\etal{\emph{et al.}}
\newcommand{\methodName}{BCCUI}

\begin{document}
\mainmatter              
\title{Breast Cancer Classification with Ultrasound Images Based on SLIC}
\titlerunning{BCCUI}  
%
\author{Zhihao Fang \and Wanyi Zhang \and He Ma(\Letter)}
\authorrunning{Zhihao Fang et al.} 
%
\tocauthor{Zhihao Fang, Wanyi Zhang and He Ma}
\institute{Sino-Dutch Biomedical and Information Engineering School, Northeastern University, Shenyang, China,\\
\email{mahe@bmie.neu.edu.cn}}

\maketitle              
\vspace*{-4pt}
\begin{abstract}
Ultrasound image diagnosis of breast tumors has been widely used in recent years. However, there are some problems of it, for instance, poor quality, intense noise and uneven echo distribution, which has created a huge obstacle to diagnosis. To overcome these problems, we propose a novel method, a breast cancer classification with ultrasound images based on SLIC (BCCUI). We first utilize the Region of Interest (ROI) extraction based on Simple Linear Iterative Clustering (SLIC) algorithm and region growing algorithm to extract the ROI at the super-pixel level. Next, the features of ROI are extracted. Furthermore, the Support Vector Machine (SVM) classifier is applied. The calculation states that the accuracy of this segment algorithm is up to 88.00\% and the sensitivity of the algorithm is up to 92.05\%, which proves that the classifier presents in this paper has certain research meaning and applied worthiness.
\end{abstract}

\section{Introduction}
\label{sec:intro}
Breast cancer is one of the most diagnosis tumor disease happened to women, which affects the health and life quality seriously. There will be approximately 268,600 cancer cases diagnosed in the United States\cite{sieg:reb}, which is around 30\% of projected cancer cases of women. Thus, how to diagnose people with breast cancer in high efficiency is one of the most challenges in the medical field. In general, there are some equipment used to screen mammary tissue, \eg mammography, magnetic resonance imaging and ultrasonography. In the clinical application, the ultrasound image is applied for further diagnosis to avoid missed diagnosis in mammography\cite{uchi:yam} and not only has the advantage of no-radiation and low-cost, but reflects the features of new blood vessels objectively in real time. However, the ultrasound image of a breast tumor is poor in quality, and the noise is serious because of its inherent imaging mechanism. Meanwhile the geometric features of the breast tumor itself are complex, and the internal echo distribution is uneven. All of these problems affect the judgment of radiologists.

In order to overcome the above problems, this paper addresses a novel method, a breast cancer classification with ultrasound images based on SLIC (BCCUI), to extract the ROI based on SLIC algorithm and classify benign and malignant breast tumor accurately and fast. Firstly, the super-pixel clustering block is obtained by using the SLIC algorithm to superimpose the denoised and sharpened breast tumor ultrasound image. Then, the region growing algorithm is used to extract the ROI. By analyzing the differences between benign and malignant features of breast tumor ultrasound images, the geometric, texture and gray features are extracted. SVM classifier is selected to classify these features in \methodName\ and get a reliable result.

The main contribution of \methodName\ is its capability of generating a relative accurate diagnosis result via common ultrasound breast images. Without deep and complicated neural network, it is more convenient for most majority of hospitals to apply this method.

The rest of this paper is organized as follows: Section~\ref{sec:2} investigates the related literature. Section~\ref{sec:3} presents the novel method in detail. Section~\ref{sec:4} demonstrates the experimental results. Section~\ref{sec:5} concludes the whole paper.

\section{Related Research}
\label{sec:2}
Computer-aided breast tumor diagnosis based on ultrasonography forms a variety of methods. In recent years, with the developing of deep learning, more and more researches try breast tumor diagnosis with neural network\cite{eht:bab,chia:huan}. However, it needs lots of hardware resources, especially, GPU resources.

The ROI extraction problem is vital for computer-aided diagnosis and the appropriate ROI determines the performance of a method directly. The greatest difficulty in this process is image segmentation, which requires to segment tumor and normal tissue. Existing image segmentation methods are mainly based on boundary detection (such as Sobel, Canny, LoG and etc), active contour model (ACM)\cite{li:chun}, threshold classification\cite{hors:kar}, snake model\cite{Chun:Hen}, watershed\cite{lewi:don}, Markov Random Field\cite{gali:ogie} and etc. They are all used to extract appropriate ROI referred to the differences of image boundaries.

One key of computer-aided diagnosis is feature extraction and there exists a large amount of methods to achieve this goal. Texture and geometry features are usually applied as the important criterion in identification of breast tumor. The present main methods are based on gray level co-occurrence matrix, Tamura texture feature, gray gradient statistics, local binary pattern, Markov random field and etc. In terms of the geometric shape of tumor, the benign tumor appears as regular shape but malignant tumor is usually irregular shape. The information commonly used is roundness, roughness, closeness and etc. Generally, these features provide the basis and reference to diagnosis.

To screening above features to figure out breast cancer, the representative features, which make classification easier, should be selected as the judgment criteria firstly. Moon \etal{} located the region of tumor, and then classified the benign and malignant tumor with interior echo and morphological characteristic\cite{moon:shen}. Uniyal \etal{} utilized RF time sequence characteristics and machine learning method to generalize the estimated probability graph of tumor and then realized classification\cite{uniy:eska}. And Nayeem \etal{} proposed a method based on sparse representation to classify tumor\cite{naye:joad}. SVM, a supervised learning method, is widely applied for classification problem, which has many advantages especially in small size sample, non-linear and high dimension problem. In this paper, we utilize SVM as the classifier.

\section{\methodName}
\label{sec:3}
\begin{figure}[H]
	\centering
	\includegraphics[width=12cm]{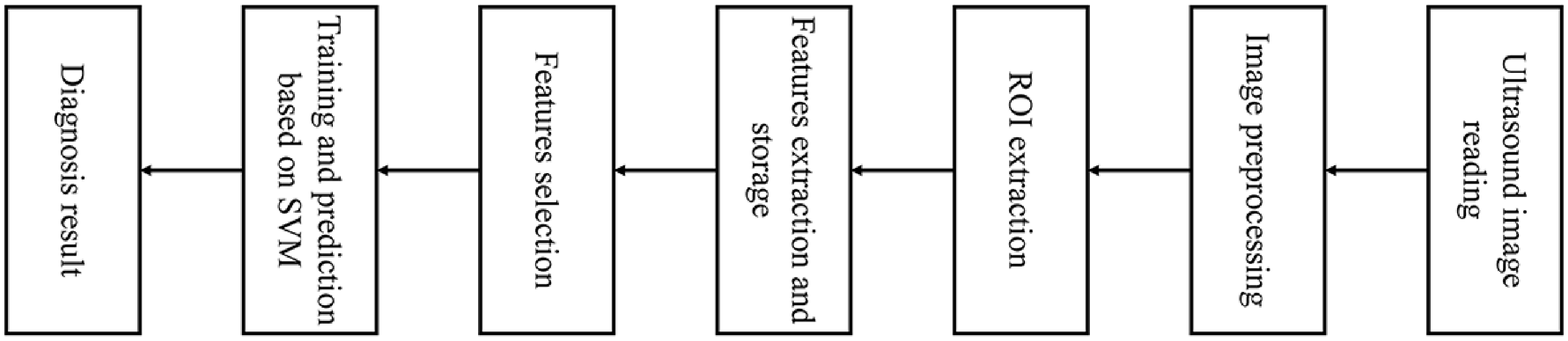}
	\caption{The whole procedure of \methodName\ for ultrasound image diagnosis.}
	\label{figure:1}
	\vspace*{-12pt}
\end{figure}

The \methodName\ can be separated as the following three parts: ROI extraction, features extraction and SVM classifier. The ultrasound image is firstly pre-processed and then handled with SLIC to extract the ROI of the breast tumor ultrasound image. Next, we extract the geometric, texture and gray features in ROI and construct the image model. After that, the features are analyzed by the model that we have trained with SVM to gain the final diagnosis result. The Figure~\ref{figure:1} illustrates whole flow diagram of the method.
\vspace*{-12pt}

\subsection{Extraction of Region of Interest}
\begin{figure}[H]
	\setlength{\fboxsep}{0.2mm}
	\subfigure[]{
		\includegraphics[width=0.45\textwidth]{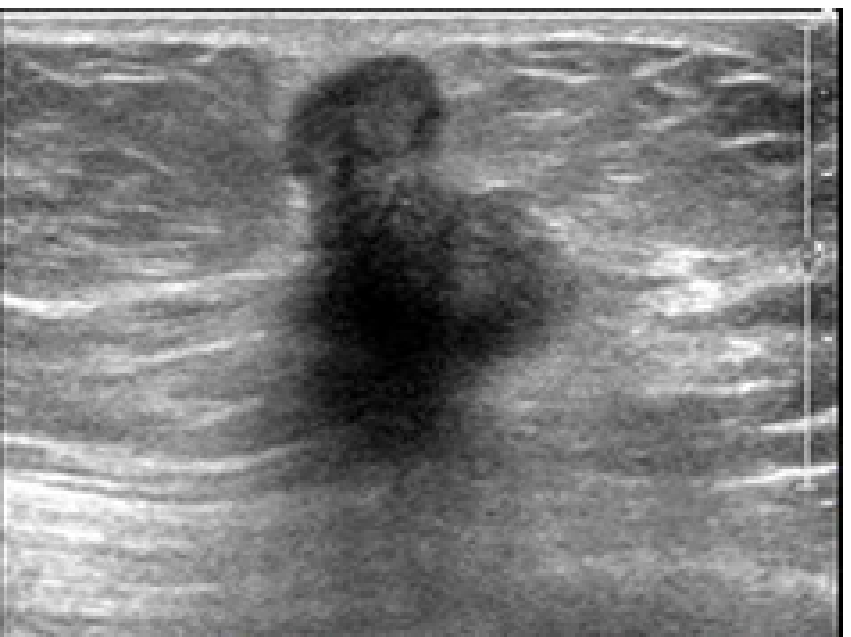}
		\label{figure:2a}
	}
	\subfigure[]{
		\hspace{-2.0mm}
		\includegraphics[width=0.45\textwidth]{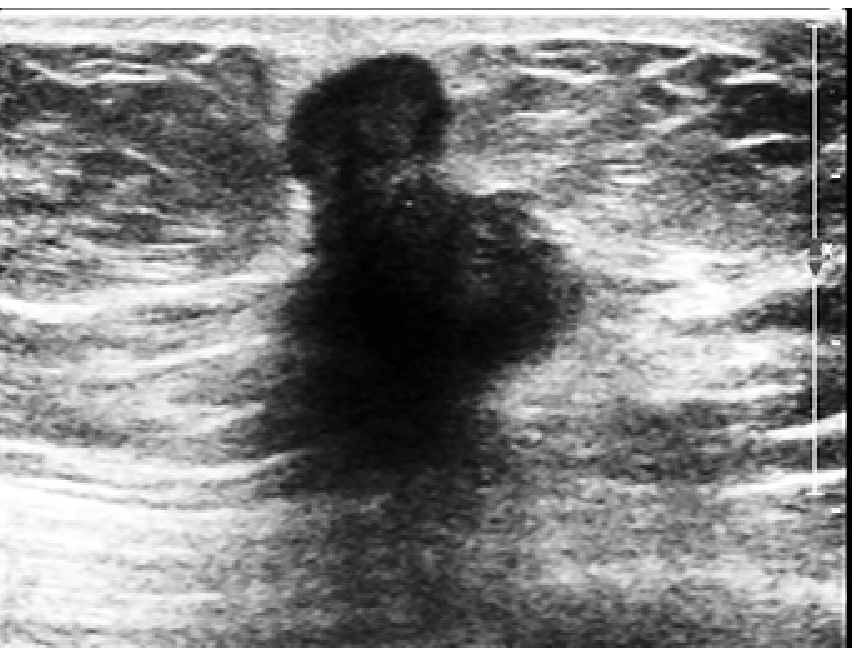}
		\label{figure:2b}
	}
	\vspace*{-10pt}
	\caption{Image preprocess. (a) is the raw ultrasound image. (b) is formed from (a) with histogram equalization and denoise process.}
	\label{figure:2}
	\vspace*{-12pt}
\end{figure}

As the ultrasound breast tumor images have the characteristic of uneven distribution of intensity and unclear boundary of tumor, histogram equalization and denoise process have been applied in original image. Figure~\ref{figure:2} shows that the contrast between tumor and background are enhanced in the processed image, which is convenient for extracting ROI.
\vspace*{-12pt}

\begin{figure}[H]
	\subfigure[]{
		\fbox{
			\includegraphics[width=0.45\textwidth]{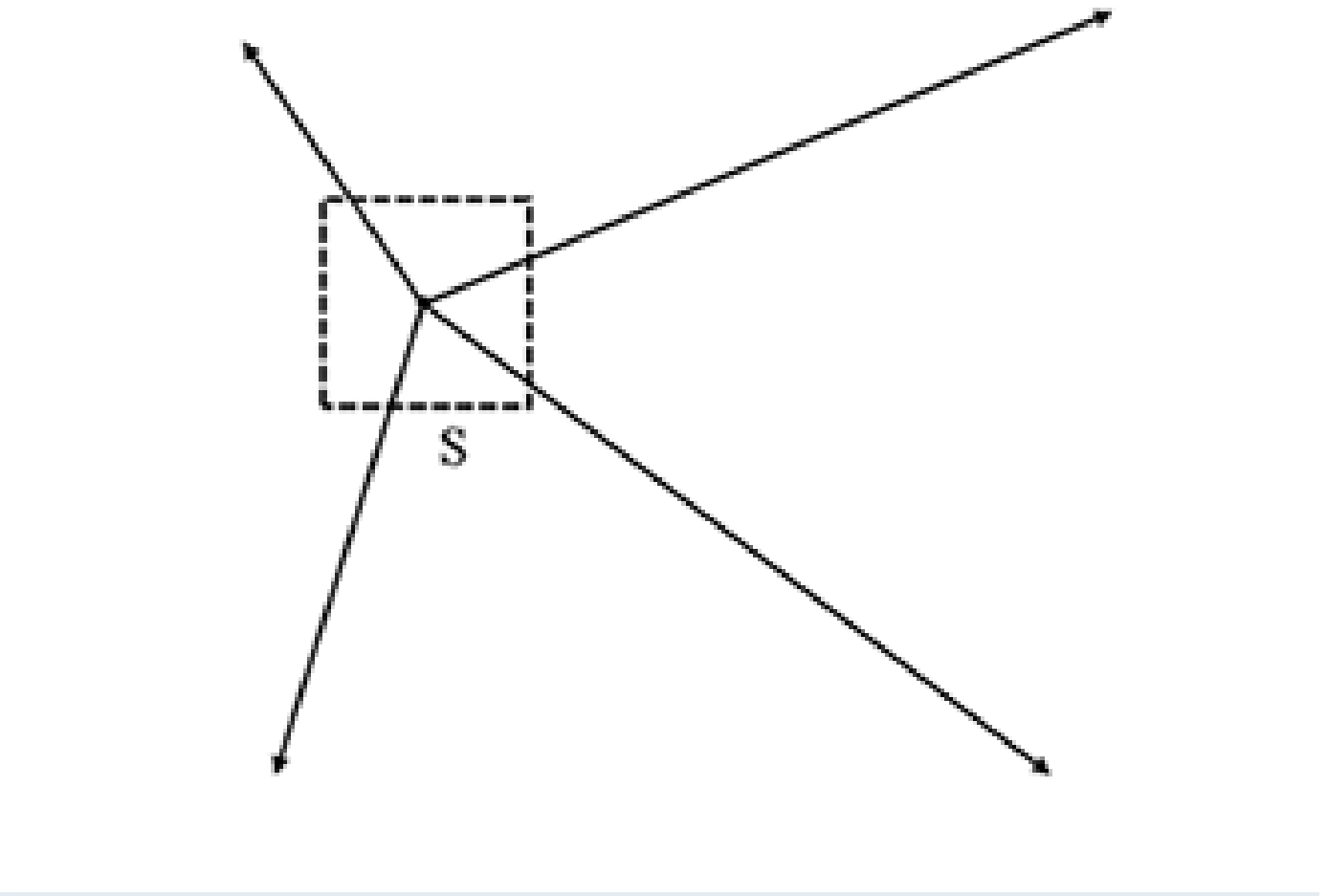}
			\label{figure:3a}}
	}
	\subfigure[]{
		\hspace{-2.0mm}
		\fbox{
			\includegraphics[width=0.45\textwidth]{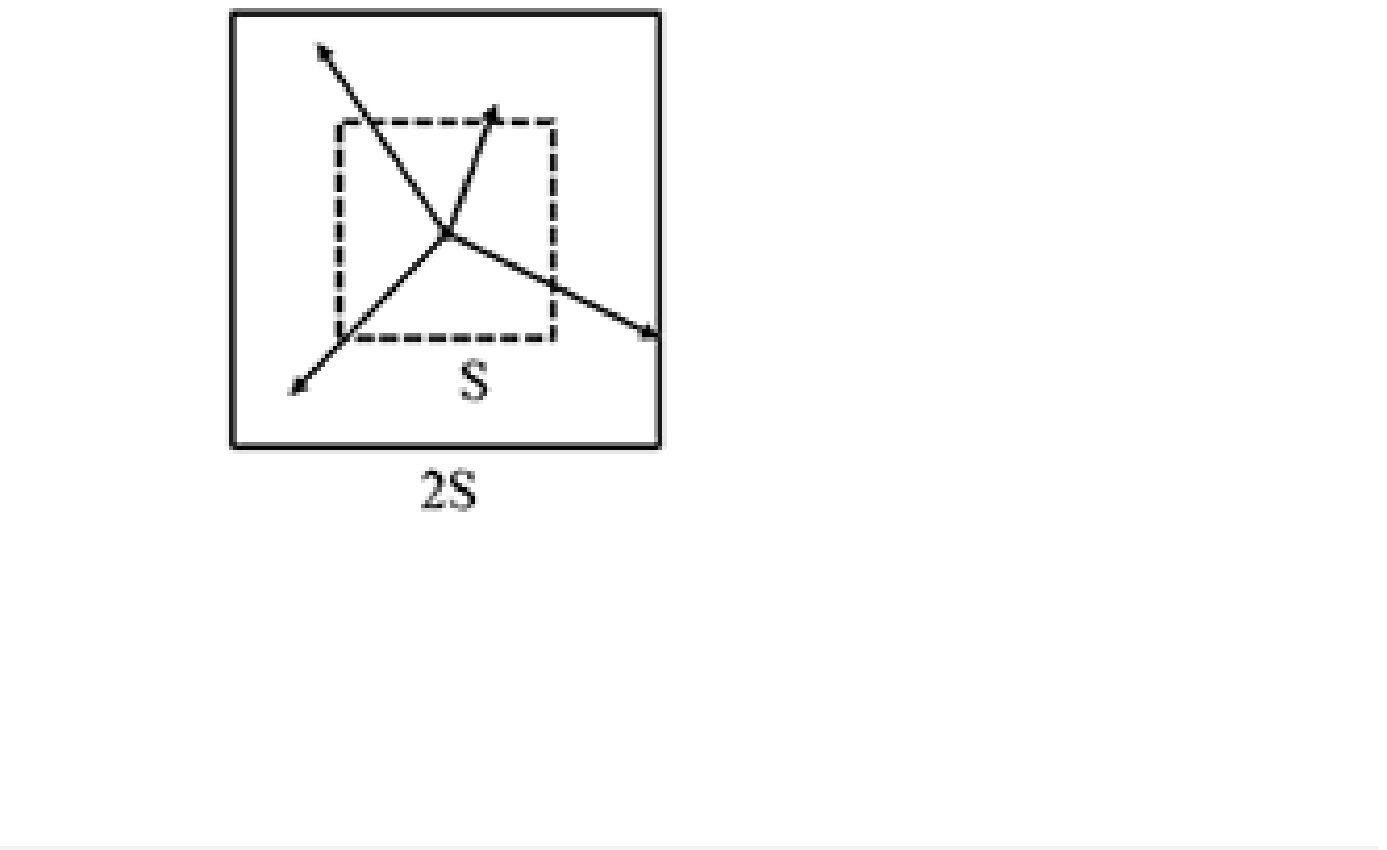}
			\label{figure:3b}}
	}
	\vspace*{-10pt}
	\caption{The comparison between K-means global search and SLIC search. (a) shows that the K-means algorithm needs to search each pixel of the image, but the SLIC only traverses the pixels in the $2S$x$2S$ area, which is faster and more effective.}
	\label{figure:3}
	\vspace*{-12pt}
\end{figure}

SLIC\cite{achc:rad} is an efficient method to decompose an image in visually homogeneous regions, which is based on the gradient ascent to segment image. Based on the color similarity and spatial distance of pixels, the super-pixel clustering block is obtained by local K-means clustering. Given the initial seed point, the step size of super-pixel is calculated as
    \begin{equation}
    S = \sqrt{\frac{N}{K}},
    \end{equation}
where $N$ is the pixel number of the ultrasound image and $K$ is the targeted number of super-pixel blocks. With the step size $S$, each block, a 3x3 area, is selected and the pixel with the lowest gradient in the block is selected as the clustering center $C_i$ which denotes the five-dimensional space as
    \begin{equation}
    C_i=[l_i,a_i,b_i,x_i,y_i],
    \end{equation}
where $[l_i,a_i,b_i]$ is the color vector in CIELAB color space and $[x_i,y_i]$ is the location of the pixel. The distance $D'$ of $C_i$ and $C_j$ can be obtained as
    \begin{equation}
    \begin{split}
    d_c&=\sqrt{(l_j-l_i)^2+(a_j-a_i)^2+(b_j-b_i)^2},\\
    d_s&=\sqrt{(x_j-x_i)^2+(y_j-y_i)^2},\\
    D'&=\sqrt{(\frac{d_c}{N_c})^2+(\frac{d_s}{N_s})^2},
    \end{split}
    \end{equation}
where $d_c$ is the color difference between the two pixels and $d_s$ is the Euclidean distance between them. $N_c$ and $N_s$, which denote color character and spatial character respectively, are normalization constant. The pixels around the seed point in $2S$x$2S$ area can be related to the seed point in the distance $D'$. Compared with K-means clustering, the search range of SLIC is limited to $2S$x$2S$ area as Figure~\ref{figure:3} shows, which makes it faster and more stable.
\vspace*{-12pt}

We combine SLIC and region-growing arithmetic to separate ROI and the fundamental step is as the following algorithm.

\floatname{algorithm}{algorithm}
\renewcommand{\algorithmicrequire}{\textbf{Input:}}
\renewcommand{\algorithmicensure}{\textbf{Output:}}
\begin{algorithm}[H]
    \caption{The extraction algorithm of ROI based on SLIC and region-growing}
    \begin{algorithmic}[1]
        \Require Ultrasound image
        \Ensure ROI boundary and segment
        \Function {GetROI}{}
        \State Read ultrasound image
        \State Get matrix of super-pixel block
        \If{The tumor has been labeled}
            \State Select the center pixel of the labeled area as seed point $(x,y)$
        \Else
            \State Select the seed point $(x,y)$ manually
        \EndIf
        \State i=0
        \For{Traverse every super-pixel block in the matrix}
            \State $g_i$=GrayAvg(i) //Calculate the average intensity of the block
            \State i++
        \EndFor
        \While{Find the appropriate neighbor}
            \For{traverse the whole SLIC matrix}
                \If{$N_4$(the super-pixel area, (x,y))==1}
                \State Save this area into list $T_0$
                \EndIf
            \EndFor
            \State i=0
            \For{traverse the list $T_0$}
                \If{($g_i$-intensity of seed point)<threshold}
                \State Save this area into list $T_1$
                \EndIf
                \State i++
            \For{traverse the pixels in list $T_1$}
                \State Record the intensity of pixels
            \EndFor
            \EndFor
            \State Copy all these pixels
        \EndWhile
        \State\Return{Save all these pixels as a new image}
        \EndFunction
    \end{algorithmic}
\end{algorithm}

\subsection{Features Extraction}
\begin{figure}[!h]
    \centering
    \begin{minipage}[t]{0.45\textwidth}
        \centering
        \includegraphics[width=2.5cm]{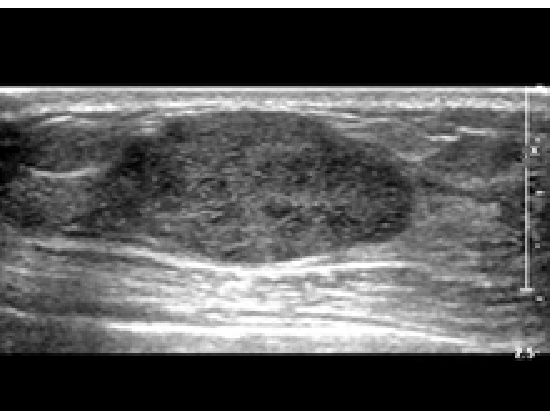}
        \includegraphics[width=2.5cm]{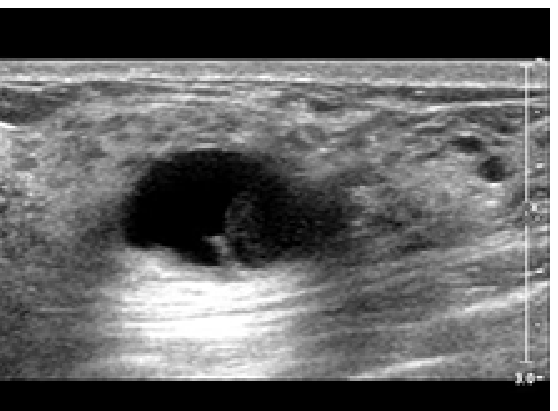}\\
        \includegraphics[width=2.5cm]{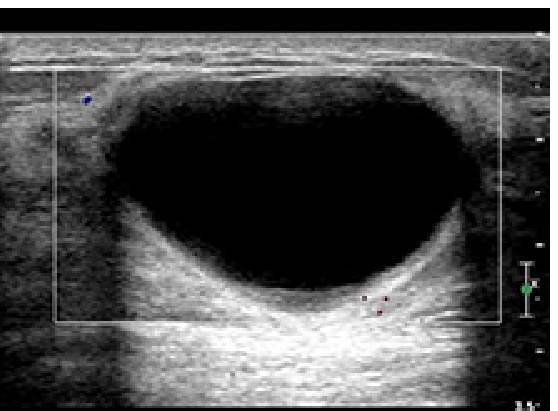}
        \includegraphics[width=2.5cm]{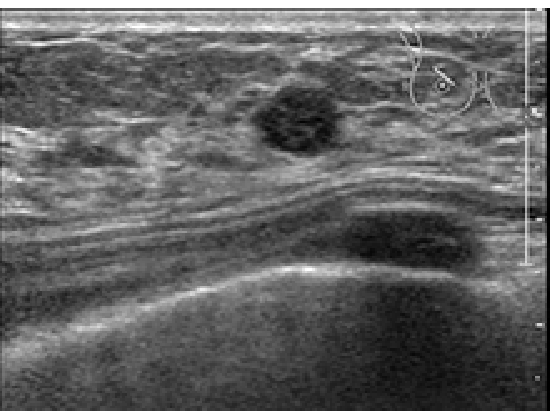}\\
        \vspace*{-10pt}
        \caption{Benign tumors. Ellipse, smooth boundary, even distribution of internal echo, posterior enhancement, smooth envelope.}
        \label{figure:4}
    \end{minipage}
    \hspace*{10pt}
    \begin{minipage}[t]{0.45\textwidth}
        \centering
        \includegraphics[width=2.5cm]{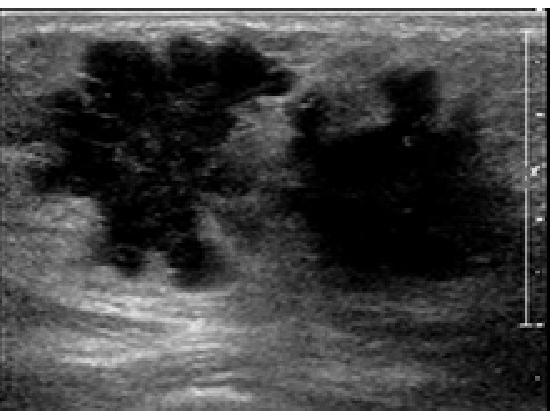}
        \includegraphics[width=2.5cm]{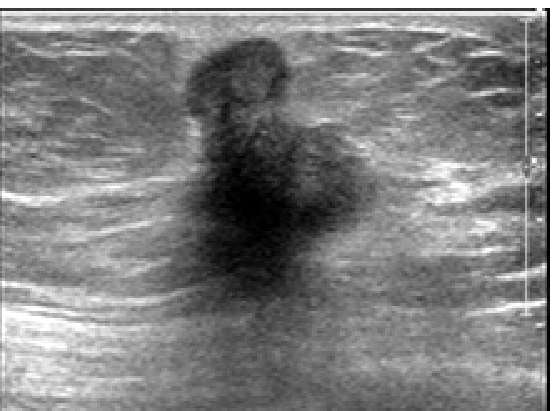}\\
        \includegraphics[width=2.5cm]{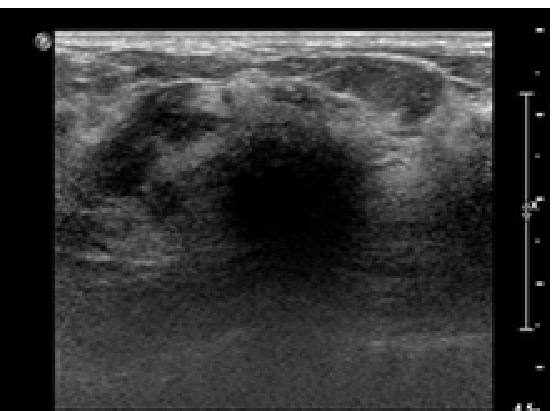}
        \includegraphics[width=2.5cm]{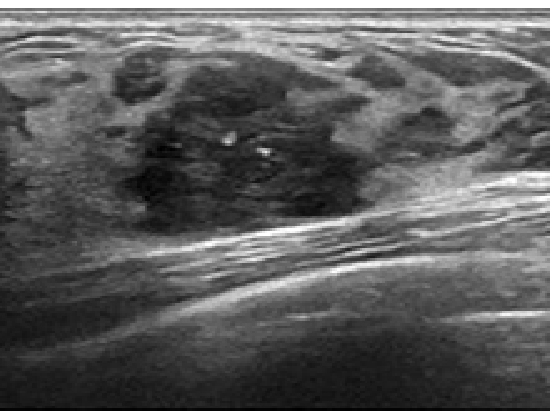}\\
        \vspace*{-10pt}
        \caption{Malignant tumor. Irregular shape, uneven boundary, posterior echo attenuation, no envelope.}
        \label{figure:5}
    \end{minipage}
\end{figure}

In the clinical diagnosis, doctors usually distinguish between benign and malignant breast tumor on the shape, texture, echo attenuation, edge character, calcification and etc\cite{dar:ruey,lang:pat,ren:malk}. The Figure~\ref{figure:4} illustrates four benign breast tumor ultrasound images and the Figure~\ref{figure:5} illustrates another four malignant tumor images. And we conclude the differences between benign and malignant breast tumor as presented in Table~\ref{table:1}.
\vspace*{-12pt}

\begin{table}[h]
    \renewcommand\arraystretch{1}
    \caption{Differences between benign and malignant breast tumor ultrasound images.}
    \vspace*{-8pt}
    \begin{tabular}[H]{p{3cm}<{\centering}p{4cm}<{\centering}p{4cm}<{\centering}}
        \hline
        character type & benign & malignant\\
        \hline
        shape       &       regular (round/ellipse)     &       irregular\\
        direction   &       parallel with epidermal     &       perpendicular to epidermal\\
        edge        &       smooth                      &       uneven\\
        boundary    &       sharp with clear echo       &       unclear \\
        internal echo&      even, low echo              &       uneven, low echo\\
        posterior echo&     increased                   &       attenuation\\
        aspect ratio&       $\leq1$                     &       $>1$\\
        micro calcification&no                          &       yes\\
        skin infiltration&  no                          &       yes\\
        \hline
    \end{tabular}
    \label{table:1}
    \vspace*{-12pt}
\end{table}

\paragraph{\textbf{Geometric Features Extraction}}
In \methodName, four geometric features are extracted. They are aspect ratio, roundness, compactness and roughness.
\begin{enumerate}
    \item \emph{Aspect Ratio}\\
    The aspect ratio can be calculated as
    \begin{equation}
        AR=\frac{h}{w}=\frac{max(i)-min(i)}{max(j)-min(j)},
    \end{equation}
    where $h$ is the height of minimum enclosing rectangle of tumor and $w$ is the width of minimum enclosing rectangle of tumor. $i$ and $j$ denote the horizontal and vertical ordinate of the rectangle's sides as the Figure~\ref{figure:6}.
    \vspace*{-10pt}
    \begin{figure}[H]
    	\subfigure[]{
    		\includegraphics[width=0.45\textwidth]{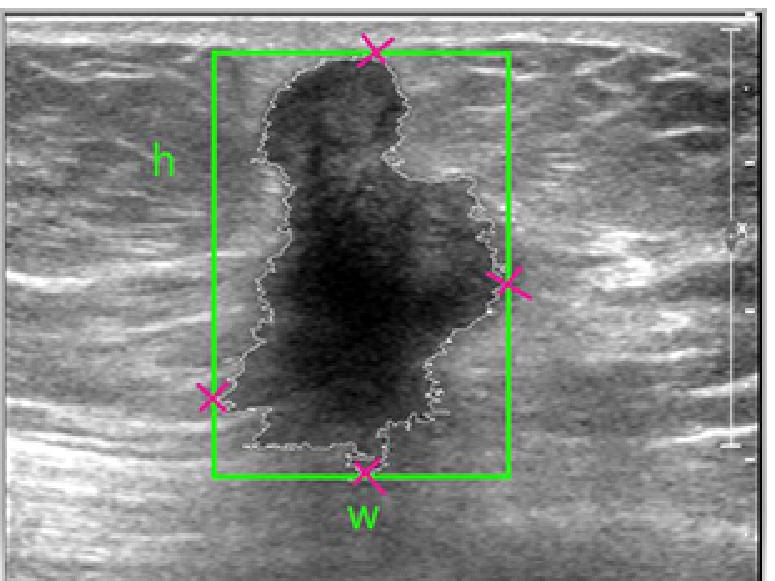}
    		\label{figure:6a}
    	}
    	\subfigure[]{
    		\hspace{-2.0mm}
    		\includegraphics[width=0.45\textwidth]{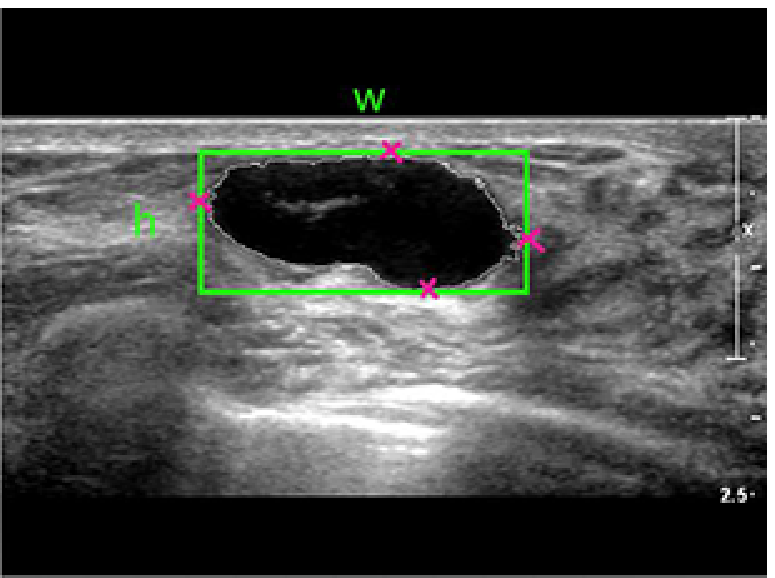}
    		\label{figure:6b}
    	}
    	\vspace*{-10pt}
    	\caption{Aspect ratio. (a) shows the height and width of malignant breast tumor and (b) shows the height and width of benign breast tumor.}
    	\label{figure:6}
    	\vspace*{-12pt}
    \end{figure}
    \item \emph{Roundness}\\
    The roundness can be calculated as
    \begin{equation}
        RD=\frac{4\pi S}{L^2},
    \end{equation}
    where $S$ is the area of the breast tumor and $L$ is perimeter. If the tumor is closer to round, the $RD$ is closer to 1.
    
    \item \emph{Compactness}\\
    The compactness can be calculated as
    \begin{equation}
        CP=\frac{S}{4\pi L^2}.
    \end{equation}
    
    \item \emph{Roughness}\\
    The Roughness can be calculated as
    \begin{equation}
        RG=\frac{1}{N}\sum_{i=1}^{N}|d(i)-d(i+1)|,
    \end{equation}
    where $N$ is pixels number of the tumor image and $d_i$ is the normalized radial length.
\end{enumerate}

\vspace*{-9pt}
\paragraph{\textbf{Texture Features Extraction}}
In \methodName, four texture features are extracted based on Gray-level Co-occurrence Matrix $GLCM(i,j|d,\theta)$. They are contrast ratio, energy, homogeneity and correlation.
\begin{enumerate}
    \item \emph{Contrast Ratio}\\
    Contrast ratio is the ratio of the luminance of the brightest pixel to that of the darkest pixel in the image, which expresses the image definition and the fluctuation of tumor groove.
    
    \item \emph{Energy}\\
    The energy can be calculated as
    \begin{equation}
        E=\sum_{i}\sum_{j}\{i,j|d,\theta\}^2.
    \end{equation}
    
    \item \emph{Homogeneity}\\
    The homogeneity can be calculated as
    \begin{equation}
        H=\sum_{i=0}^{L-1}\sum_{j=0}{L-1}\frac{p(i,j|d,\theta)}{1+(i-j)^2}.
    \end{equation}
    
    \item \emph{Correlation}\\
    The correlation can be calculated as
    \begin{equation}
        COR=\sum_{i}\sum_{j}\frac{(i-\mu_x)(j-\mu_y)P(i,j|d,\theta)}{\sigma_x\sigma_y},
    \end{equation}
    where $\mu_x$, $\mu_y$, $\sigma_x$ and $\sigma_y$ are the mean values and standard deviations of the normalized $GLCM$ respectively as
    \begin{equation}
    \begin{split}
        \mu_x&=\sum_{i}i\sum_{j}P(i,j|d,\theta),
        \mu_y=\sum_{j}j\sum_{i}P(i,j|d,\theta),\\
        \sigma_x&=\sum_i(i-\mu_x)^2\sum_j P(i,j|d,\theta),
        \sigma_y=\sum_j(j-\mu_y)^2\sum_j P(i,j|d,\theta).
    \end{split}
    \end{equation}
\end{enumerate}

\begin{figure}
	\subfigure[]{
		\includegraphics[width=0.45\textwidth]{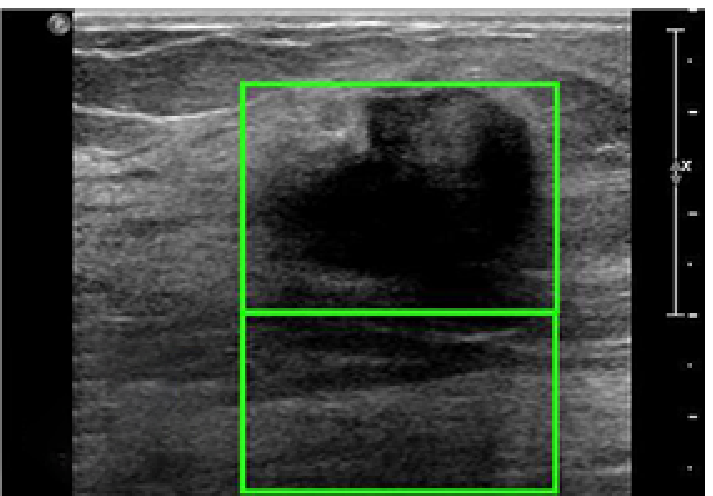}
	}
	\subfigure[]{
		\hspace{-2.0mm}
		\includegraphics[width=0.45\textwidth]{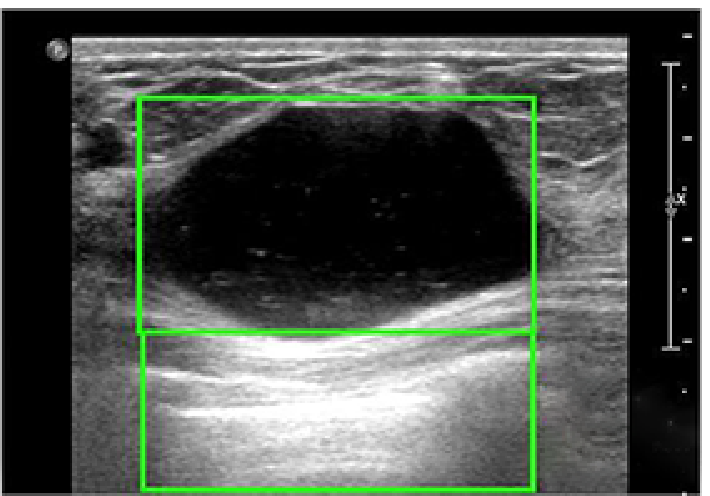}
	}
	\vspace*{-10pt}
	\caption{Posterior echo. The first rectangle in image marks the location of tumor and the second rectangle marks the posterior echo. (a) is the image of malignant tumor and (b) is the image of benign tumor.}
	\label{figure:7}
\end{figure}

\vspace*{-9pt}
\paragraph{\textbf{Gray Feature Extraction}}
The posterior echo attenuation of the tumor is also an important basis for diagnosis. Thus we take the gray mean of the area at the back of tumor as a feature. As the Figure~\ref{figure:7} illustrates, if there's echo attenuation behind the tumor, the probability of malignant breast tumor is higher. If there's no clear echo attenuation, the probability of malignant breast tumor is lower.
\vspace*{-12pt}
\vspace*{-12pt}

In \methodName, the attenuation coefficient represents the gray feature as
\begin{equation}
    AC=\frac{Avg_{ROI}}{Avg_{back}}
\end{equation}
where $Avg_{ROI}$ is the mean intensity value of the tumor area and $Avg_{back}$ is the mean intensity value of the rectangle.

\subsection{Classification}
Support-vector machines\cite{ben:asa}, a supervised learning models with associated learning algorithms that analyze data, is widely used for classification and regression analysis. SVM can efficiently perform a non-linear classification using kernel trick which maps the input data into high dimensional space. We recommend using RBF as the kernel which is
\begin{equation}
    K(x,x_i)=exp(-\frac{||x-x_i||^2}{\delta^2}),
\end{equation}
And we find that using Sigmoid function also obtains the satisfied result of classification. 

\section{Experimental Results}
\label{sec:4}
\begin{figure}
	\subfigure[]{
		\includegraphics[width=0.45\textwidth]{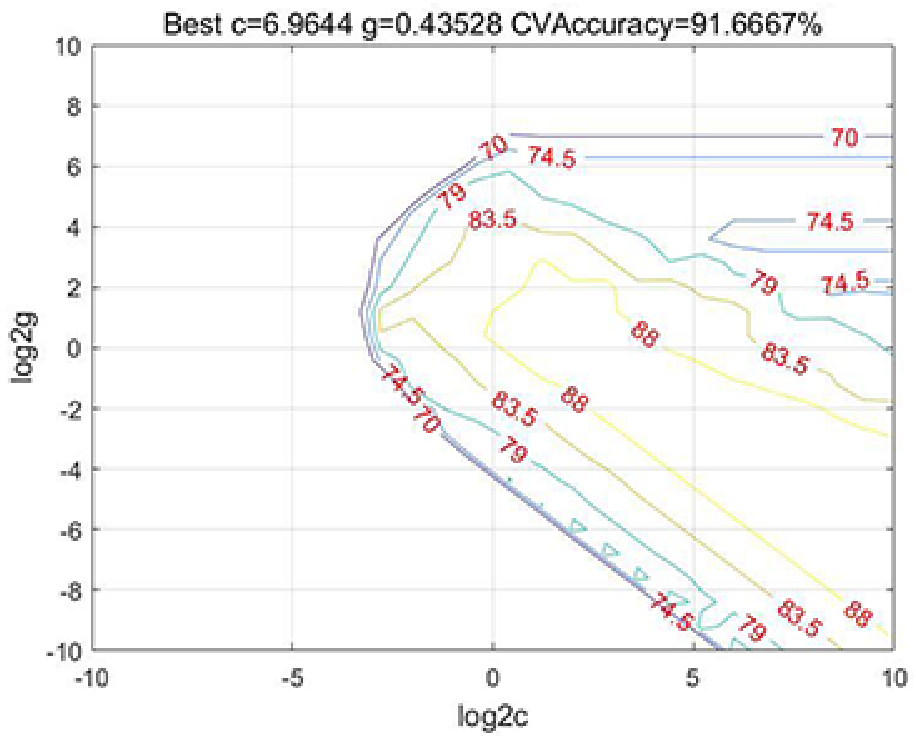}
	}
	\subfigure[]{
		\hspace{-2.0mm}
		\includegraphics[width=0.45\textwidth]{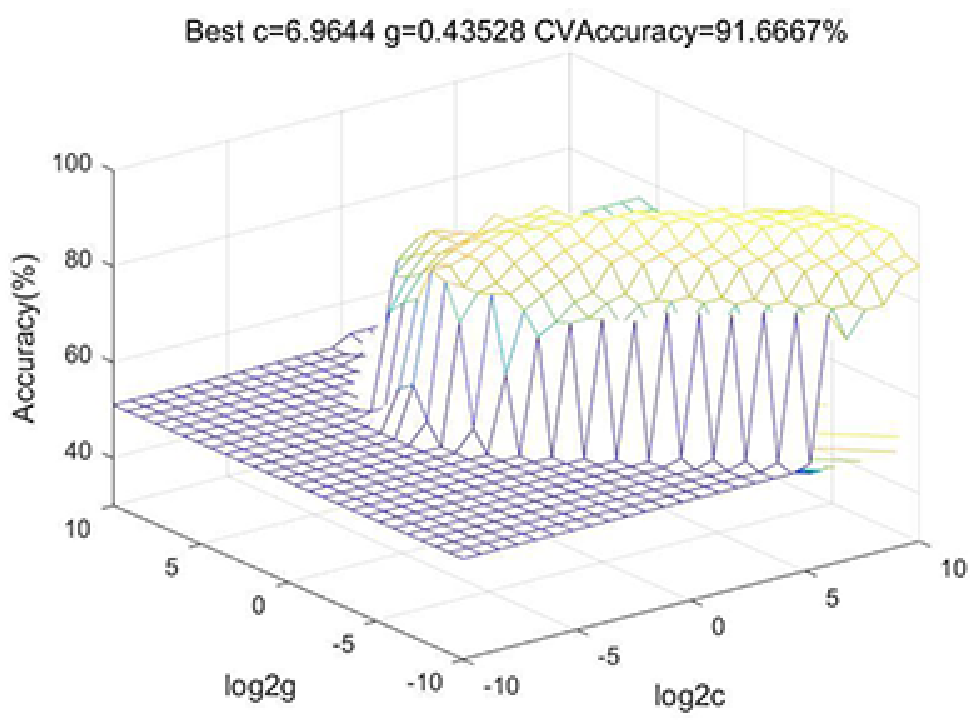}
	}
	\vspace*{-10pt}
	\caption{Result of parameter optimization. Best $c=6.9644$, $g=0.43528$ and $CVAcurracy=91.6667\%$. (a) is the two-dimensional diagram of result. (b) is the three-dimensional diagram of result.}
	\label{figure:9}
	\vspace*{-20pt}
\end{figure}

\begin{figure}
    \centering
	\includegraphics[width=0.45\textwidth]{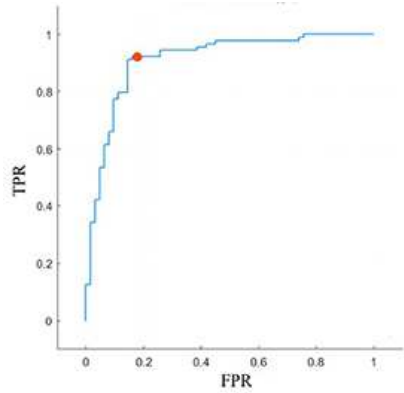}
	\vspace*{-12pt}
	\caption{ROC curve of SVM.}
	\label{figure:10}
	\vspace*{-12pt}
\end{figure}

\begin{figure}
	\newcommand{\pwidth}{3cm}
	\newcommand{\hwidth}{0.1cm}
	\newcommand{\width}{3cm}
	\begin{tabular}{p{\pwidth}<{\centering}p{\pwidth}<{\centering}p{\pwidth}<{\centering}p{\pwidth}<{\centering}}
		\footnotesize Raw image & ROI contour & Extraction result (binary image) & Extraction result (gray image)\vspace{\hwidth}\\
		\begin{minipage}{\width}
			\includegraphics[width=\width]{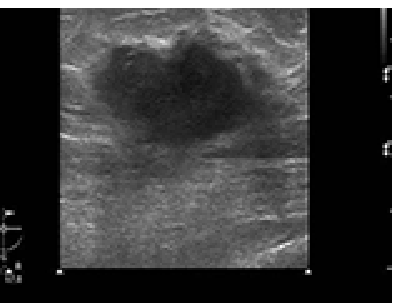}
		\end{minipage} &
		\begin{minipage}{\width}
			\includegraphics[width=\width]{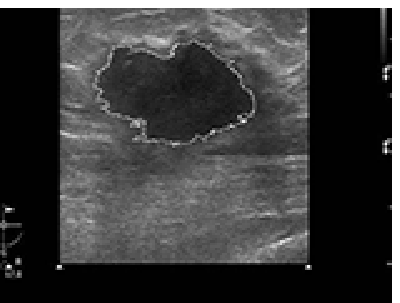}
		\end{minipage} &
		\begin{minipage}{\width}
			\includegraphics[width=\width]{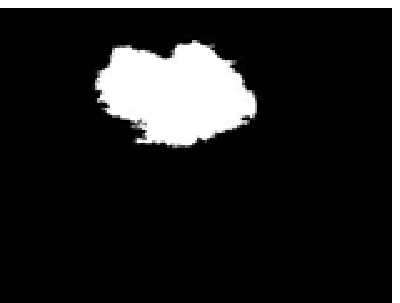}
		\end{minipage} &
		\begin{minipage}{\width}
			\includegraphics[width=\width]{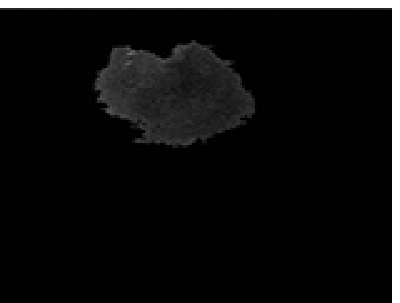}
		\end{minipage}\vspace{\hwidth}\\
		\begin{minipage}{\width}
			\includegraphics[width=\width]{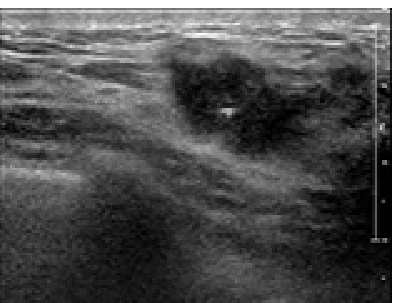}
		\end{minipage} &
		\begin{minipage}{\width}
			\includegraphics[width=\width]{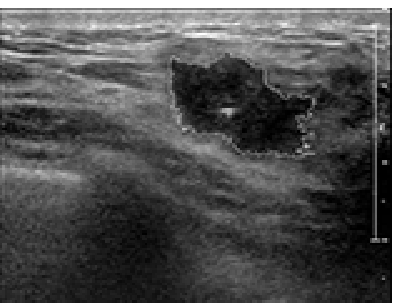}
		\end{minipage} &
		\begin{minipage}{\width}
			\includegraphics[width=\width]{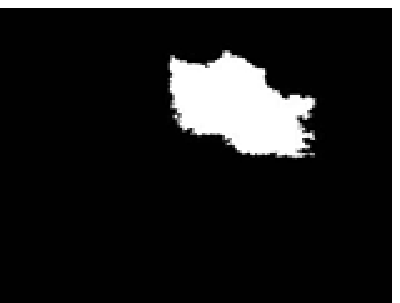}
		\end{minipage} &
		\begin{minipage}{\width}
			\includegraphics[width=\width]{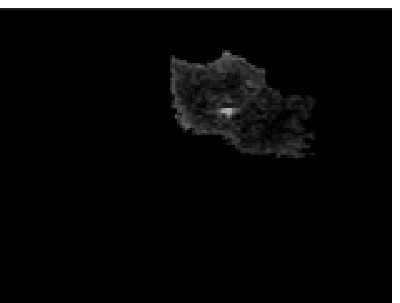}
		\end{minipage}\vspace{\hwidth}\\
		\begin{minipage}{\width}
			\includegraphics[width=\width]{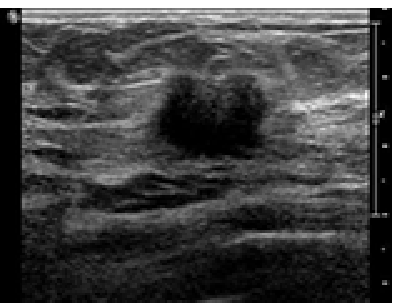}
		\end{minipage} &
		\begin{minipage}{\width}
			\includegraphics[width=\width]{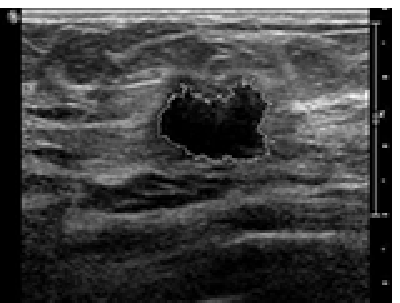}
		\end{minipage} &
		\begin{minipage}{\width}
			\includegraphics[width=\width]{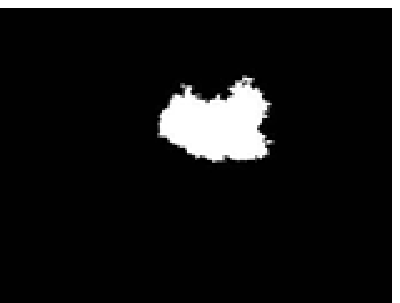}
		\end{minipage} &
		\begin{minipage}{\width}
			\includegraphics[width=\width]{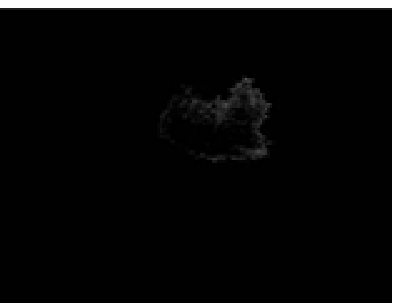}
		\end{minipage}\vspace{\hwidth}\\
		\begin{minipage}{\width}
			\includegraphics[width=\width]{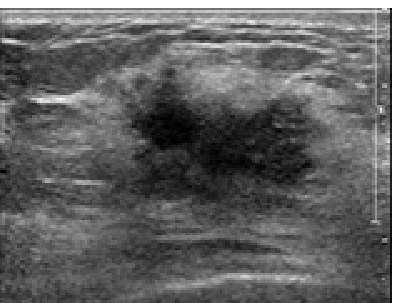}
		\end{minipage} &
		\begin{minipage}{\width}
			\includegraphics[width=\width]{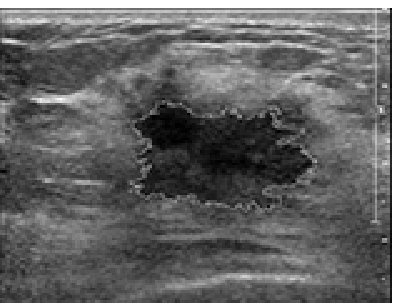}
		\end{minipage} &
		\begin{minipage}{\width}
			\includegraphics[width=\width]{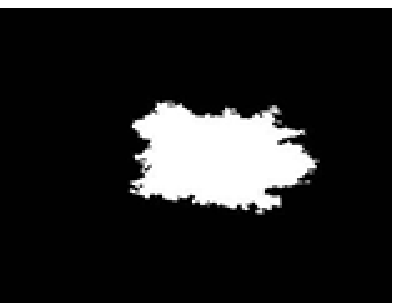}
		\end{minipage} &
		\begin{minipage}{\width}
			\includegraphics[width=\width]{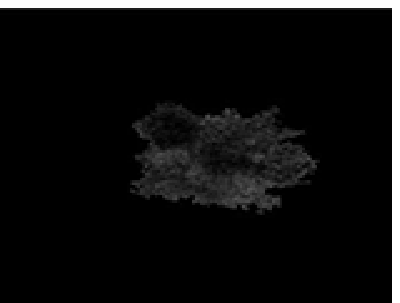}
		\end{minipage}\vspace{\hwidth}\\
		\begin{minipage}{\width}
			\includegraphics[width=\width]{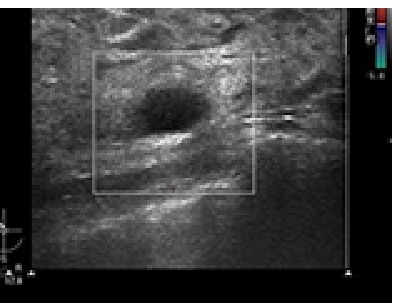}
		\end{minipage} &
		\begin{minipage}{\width}
			\includegraphics[width=\width]{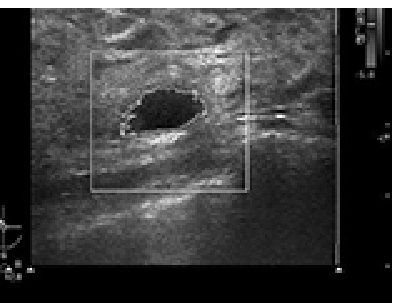}
		\end{minipage} &
		\begin{minipage}{\width}
			\includegraphics[width=\width]{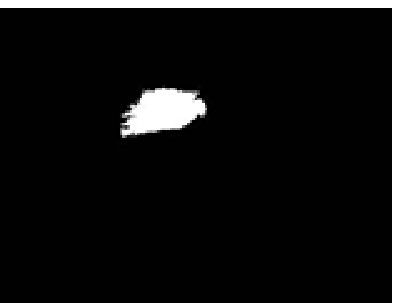}
		\end{minipage} &
		\begin{minipage}{\width}
			\includegraphics[width=\width]{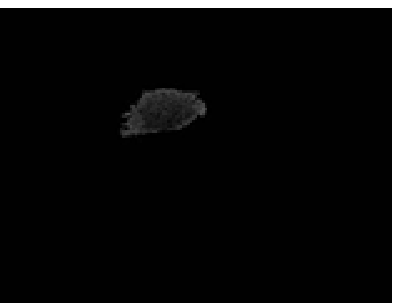}
		\end{minipage}\vspace{\hwidth}\\
		\begin{minipage}{\width}
			\includegraphics[width=\width]{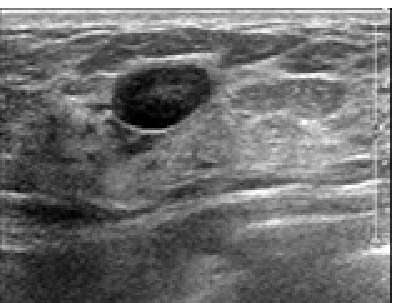}
		\end{minipage} &
		\begin{minipage}{\width}
			\includegraphics[width=\width]{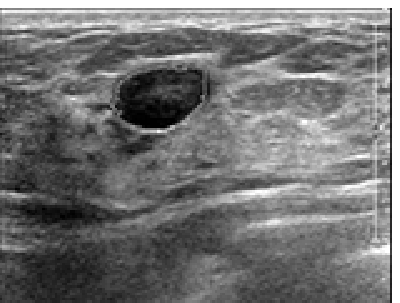}
		\end{minipage} &
		\begin{minipage}{\width}
			\includegraphics[width=\width]{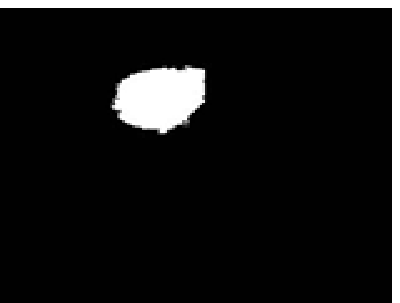}
		\end{minipage} &
		\begin{minipage}{\width}
			\includegraphics[width=\width]{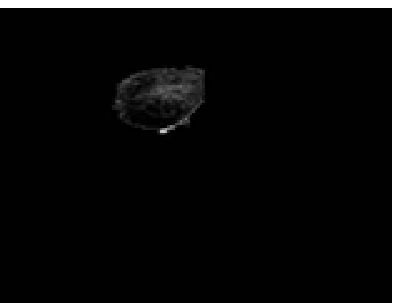}
		\end{minipage}\vspace{\hwidth}\\
		\begin{minipage}{\width}
			\includegraphics[width=\width]{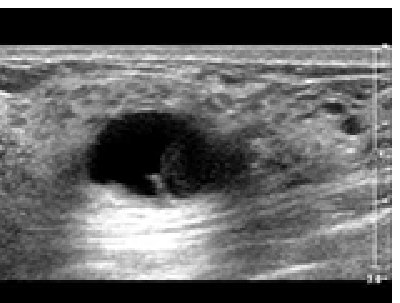}
		\end{minipage} &
		\begin{minipage}{\width}
			\includegraphics[width=\width]{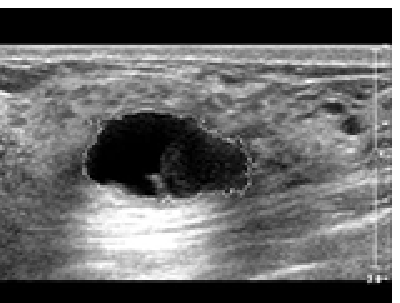}
		\end{minipage} &
		\begin{minipage}{\width}
			\includegraphics[width=\width]{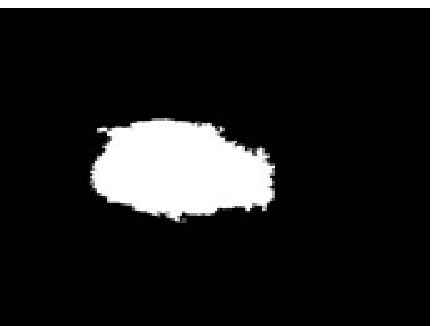}
		\end{minipage} &
		\begin{minipage}{\width}
			\includegraphics[width=\width]{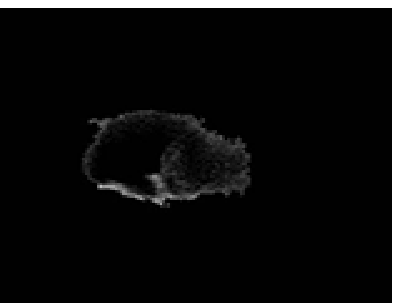}
		\end{minipage}\vspace{\hwidth}\\
		\begin{minipage}{\width}
			\includegraphics[width=\width]{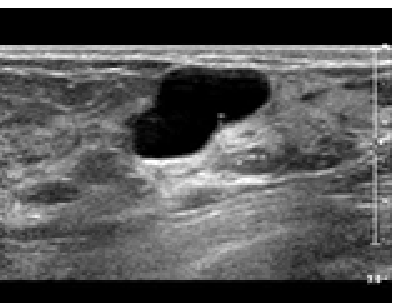}
		\end{minipage} &
		\begin{minipage}{\width}
			\includegraphics[width=\width]{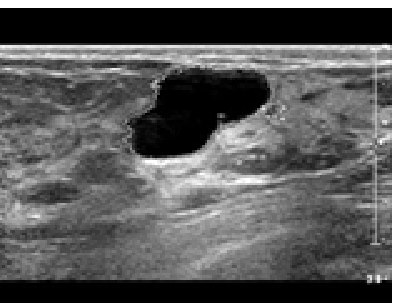}
		\end{minipage} &
		\begin{minipage}{\width}
			\includegraphics[width=\width]{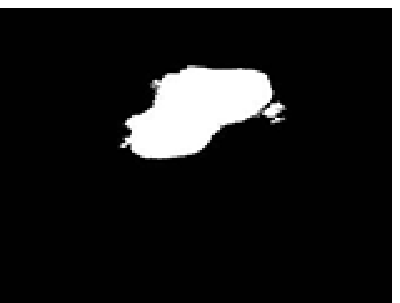}
		\end{minipage} &
		\begin{minipage}{\width}
			\includegraphics[width=\width]{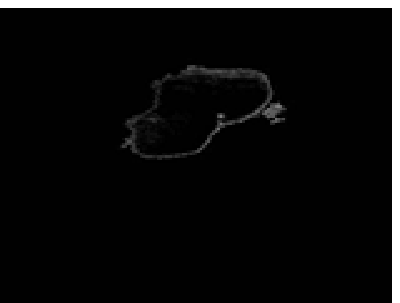}
		\end{minipage}
	\end{tabular}
	\vspace{0.3cm}
	\caption{Some ROI extraction results.}
	\label{figure:8}
\end{figure}

The breast ultrasounds image dataset of our experiment came from imaging department of local hospital, which contains the breast tumor ultrasound images of 150 patients. All cases were confirmed by operation and pathology. There are 88 malignant breast tumor cases and 62 benign tumor cases. The size of all images is 580x775 and the format is PNG. We compile our code by using MATLAB R2017a and run it with a 2.8GHz Intel Core i5-8400 CPU.
\vspace*{-9pt}
\paragraph{\textbf{ROI Extraction}} In Figure~\ref{figure:8}, we exhibit some ROI extraction results. The former four groups are benign and the later four groups are malignant. Observed Figure~\ref{figure:8}, \methodName\ divides the tumors from image clearly with the number of super-pixel blocks as 50.
\vspace*{-9pt}
\paragraph{\textbf{Parameter Optimization}} We take LIBSVM toolkit to find the best parameter of the classifier. In SVM, the penalty parameter $c$ and the radius of RBF $g$ affect the classification result directly, thus, we utilize SVMcgForClass function to search the best parameter $c$ and $g$ in grid. As the Figure~\ref{figure:9} illustrates, the best condition is when $c=6.9644$ and $g=0.43528$.
\vspace*{-12pt}
\vspace*{-9pt}
\paragraph{\textbf{Classification}}
Using the above optimized parameters to training the model with our dataset, the result is obtained as Table~\ref{table:2}, where TP, TN, FP and FN denote true positives, true negatives, false negatives and false positives respectively. Note that we have used 5-fold cross validation in our experiment.
\vspace*{-18pt}
\begin{table}[H]
    \renewcommand\arraystretch{1}
    \caption{Result of 5-fold cross validation.}
    \vspace*{-4pt}
    \begin{tabular}[H]{p{1.5cm}<{\centering}p{2.5cm}<{\centering}p{2.5cm}<{\centering}p{2.5cm}<{\centering}p{2.5cm}<{\centering}}
        \hline
        Index & TP & TN & FP & FN\\
        \hline
        1   &   20  &   9   &   1   &   0\\
        2   &   16  &   12  &   0   &   2\\
        3   &   14  &   11  &   4   &   1\\
        4   &   18  &   9   &   3   &   0\\
        5   &   13  &   10  &   3   &   4\\
        \hline
        total&  81  &   51  &   11  &   7\\
        \hline
    \end{tabular}
    \label{table:2}
\end{table}
\vspace*{-12pt}
Given TP, TN, FP and FN, some evaluation index is calculated as Table~\ref{table:3}.
\begin{table}
    \renewcommand\arraystretch{1}
    \vspace*{-8pt}
    \caption{Evaluation of the classification.}
    \vspace*{-4pt}
    \begin{tabular}[H]{p{3cm}<{\centering}p{6cm}<{\centering}p{2.5cm}<{\centering}}
        \hline
        Evaluation index & Formula & Value\\
        \hline
        Accuracy            &   $(TP+TN)/(TP+TN+FP+FN)$    &   88.00\%\\
        Sensibility         &   $TP/(TP+FN)$               &   92.05\%\\
        Specificity         &   $TN/(TN+FP)$               &   82.26\%\\
        Positive Accuracy   &   $TP/(TP+FP)$               &   88.04\%\\
        Negative Accuracy   &   $TN/(TN+FN)$               &   87.93\%\\
        \hline
    \end{tabular}
    \label{table:3}
    \vspace*{-12pt}
\end{table}
\vspace*{-9pt}
\paragraph{\textbf{Receiver Operating Characteristic Curve}} Figure~\ref{figure:10} shows the ROC curve of the experiment. At the red dot, the sensibility and specificity are higher simultaneously in Figure~\ref{figure:10}. And we calculate the AUC is 0.91, which proves the classifier and \methodName\ is effective and have a good result.
\vspace*{-16pt}

\section{Conclusion}
\label{sec:5}
This paper introduced a novel method to classify benign and malignant breast tumor with raw ultrasound images. It has advantages of fast, accurate, friendly to embedded device and etc. From a series of experiment, it is proven that \methodName\ will have a widely applied prospect.

\begin{thebibliography}{10}

\bibitem{sieg:reb}
Siegel, Rebecca L. and Miller, Kimberly D. and Jemal, Ahmedin, Cancer statistics, 2019, CA: A Cancer Journal for Clinicians. 69, 7-34 (2019). \url{doi:10.3322/caac.21551}
\bibitem{uchi:yam}
Ken Uchida, Akinori Yamashita, Kazumi Kawase, Kentarou Kamiya, Screening ultrasonography revealed 15\% of mammographically occult breast cancers, Breast Cancer. 15, 165 (2008). \url{doi:10.1007/s12282-007-0024-x}
\bibitem{eht:bab}
Ehteshami Bejnordi, Babak and Veta, Mitko and Johannes van Diest, Paul and van Ginneken, Bram and Karssemeijer, Nico and Litjens, Geert and van der Laak, Jeroen A. W. M. and and the CAMELYON16 Consortium, Diagnostic Assessment of Deep Learning Algorithms for Detection of Lymph Node Metastases in Women With Breast Cancer, JAMA, 318-22, 2199-2210 (2017), \url{doi:10.1001/jama.2017.14585}
\bibitem{chia:huan}
T. Chiang, Y. Huang, R. Chen, C. Huang and R. Chang, Tumor Detection in Automated Breast Ultrasound Using 3-D CNN and Prioritized Candidate Aggregation, IEEE Transactions on Medical Imaging, 38-1, 240-249 (2019), \url{doi:10.1109/TMI.2018.2860257}
\bibitem{li:chun}
Li Wang, Chunming Li, Quansen Sun, Deshen Xia, Chiu-Yen Kao, Active contours driven by local and global intensity fitting energy with application to brain MR image segmentation. Computerized Medical Imaging and Graphics. 33-7, 520-531 (2009). \url{doi:10.1016/j.compmedimag.2009.04.010}
\bibitem{hors:kar}
Horsch, Karla and Giger, Maryellen L. and Venta, Luz A. and Vyborny, Carl J, Computerized diagnosis of breast lesions on ultrasound, Medical Physics. 29-2, 157-164 (2002). \url{doi:10.1118/1.1429239}
\bibitem{Chun:Hen}
Chung-Ming Chen, Henry Horng-Shing Lu, Yu-Chen Lin, An early vision-based snake model for ultrasound image segmentation, Ultrasound in Medicine and Biology. 2-2, 273-285 (2000). \url{doi:10.1016/S0301-5629(99)00140-4}
\bibitem{lewi:don}
S. H. Lewis and A. Dong, Detection of breast tumor candidates using marker-controlled watershed segmentation and morphological analysis, 2012 IEEE Southwest Symposium on Image Analysis and Interpretation, 1-4 (2012). \url{doi:10.1109/SSIAI.2012.6202438}
\bibitem{gali:ogie}
Galińska M., Ogiegło W., Wijata A., Juszczyk J., Czajkowska J, Breast Cancer Segmentation Method in Ultrasound Images, Innovations in Biomedical Engineering, 23-31 (2018). \url{doi:10.1007/978-3-319-70063-2_3}
\bibitem{moon:shen}
W. K. Moon, Y. Shen, M. S. Bae, C. Huang, J. Chen and R. Chang, Computer-Aided Tumor Detection Based on Multi-Scale Blob Detection Algorithm in Automated Breast Ultrasound Images, IEEE Transactions on Medical Imaging, 32, 7 (2013). \url{doi:10.1109/TMI.2012.2230403}
\bibitem{uniy:eska}
N. Uniyal, H. Eskandari, P. Abolmaesumi, S. Sojoudi, P. Gordon, L. Warren, R. N. Rohling, S. E. Salcudean and M. Moradi, Ultrasound RF Time Series for Classification of Breast Lesions, IEEE Transactions on Medical Imaging, 34-2, 652-661 (2015). \url{doi:10.1109/TMI.2014.2365030}
\bibitem{naye:joad}
M. A. R. Nayeem, M. A. M. Joadder, S. A. Shetu, F. R. Jamil and A. A. Helal, Feature selection for breast cancer detection from ultrasound images, International Conference on Informatics, Electronics \& Vision, 1-6 (2014), \url{doi:10.1109/ICIEV.2014.6850813}
\bibitem{achc:rad}
Achanta, Radhakrishna and Shaji, Appu and Smith, Kevin and  Lucchi, Aurélien and Fua, Pascal and Süsstrunk, Sabine, SLIC Superpixels, Technical report, (2015).
\bibitem{dar:ruey}
Dar-Ren Chen, Ruey-Feng Chang, Yu-Len Huang, Breast cancer diagnosis using self-organizing map for sonography, Ultrasound in Medicine and Biology, 26, 3 (2000), \url{doi:10.1016/S0301-5629(99)00156-8}
\bibitem{lang:pat}
Langley, Pat and Sage, Stephanie, Induction of Selective Bayesian Classifiers, Proceedings of the Tenth International Conference on Uncertainty in Artificial Intelligence, 399-406 (1994).
\bibitem{ren:malk}
Ren and Malik, Learning a classification model for segmentation, Proceedings Ninth IEEE International Conference on Computer Vision, 1, 10-17 (2003), \url{doi:10.1109/ICCV.2003.1238308}
\bibitem{ben:asa}
Ben-Hur, Asa and Horn, David and Siegelmann, Hava T. and Vapnik, Vladimir, Support vector clustering, Journal of Machine Learning Research, 2, 125-137 (2001)

\end{thebibliography}
\end{document}